\documentclass{article}
\usepackage{spconf,amsmath,graphicx}
\usepackage{amsmath}
\usepackage{amssymb}
\usepackage{enumitem}
\setitemize{noitemsep,topsep=0pt,parsep=0pt,partopsep=0pt}

\title{LONG-TAILED FOOD CLASSIFICATION}
\name{Jiangpeng He$^*$ \quad Luotao Lin$^{\dagger}$ \quad Heather Eicher-Miller$^{\dagger}$ \quad Fengqing Zhu$^*$ }

\address{$^*$School of Electrical and Computer Engineering \quad $^{\dagger}$Department of Nutrition Science \\ \\
Purdue University, West Lafayette, Indiana, U.S.A}

\begin{document}
%
\maketitle
\begin{abstract}
Food classification serves as the basic step of image-based dietary assessment to predict the types of foods in each input image. However, food image predictions in a real world scenario are usually long-tail distributed among different food classes, which cause heavy class-imbalance problems and a restricted performance. In addition, none of the existing long-tailed classification methods focus on food data, which can be more challenging due to the lower inter-class and higher intra-class similarity among foods. In this work, we first introduce two new benchmark datasets for long-tailed food classification including Food101-LT and VFN-LT where the number of samples in VFN-LT exhibits the real world long-tailed food distribution. Then we propose a novel 2-Phase framework to address the problem of class-imbalance by (1) undersampling the head classes to remove redundant samples along with maintaining the learned information through knowledge distillation, and (2) oversampling the tail classes by performing visual-aware data augmentation. We show the effectiveness of our method by comparing with existing state-of-the-art long-tailed classification methods and show improved performance on both Food101-LT and VFN-LT benchmarks. The results demonstrate the potential to apply our method to  related real life applications.
\end{abstract}
\begin{keywords}
Image Classification, Long-tailed Distribution, Deep Learning, Knowledge Distillation
\end{keywords}

\begin{figure}[t]
\begin{center}
  \includegraphics[width=1.\linewidth]{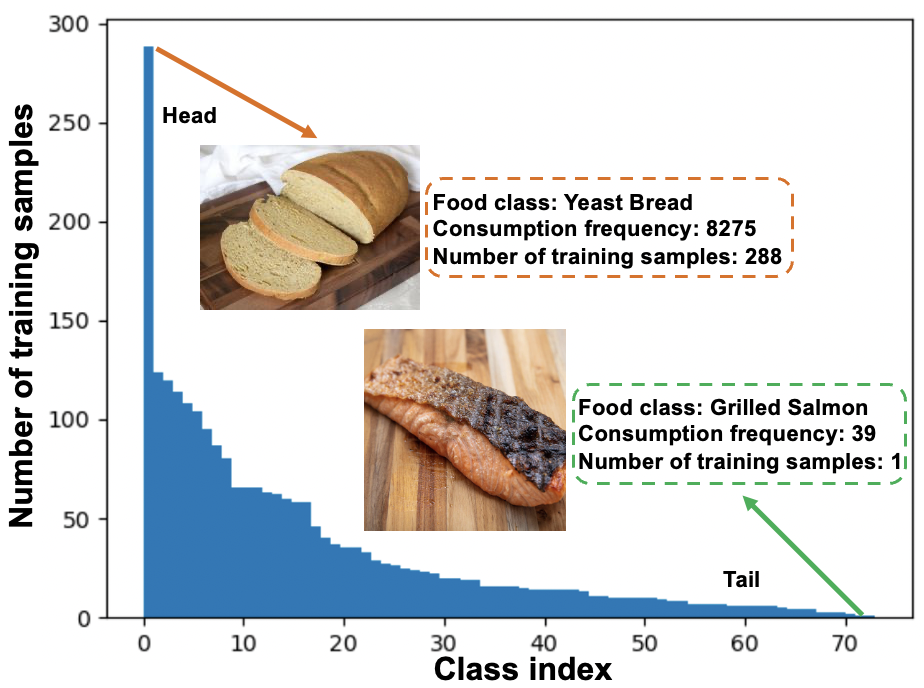}
  \caption{The overview of the VFN-LT that exhibits real-world long-tailed food distribution. The number of training samples are assigned based on consumption frequency, which is matched through NHANES from 2009–2016 among 17,796 U.S. healthy adults. }
  \label{fig:vfn-lt}
\end{center}
\end{figure}

\section{Introduction}
\label{sec:intro}
Accurate identification of food is critical to image-based dietary assessment~\cite{he2020multitask,he2021end}, which facilitates matching the food to the proper identification of that food in a nutrient database with corresponding nutrient composition~\cite{shao2021_nutri_database}. Such linkage makes it possible to determine dietary links to health and disease, such as diabetes. Dietary assessment, therefore, is very important to health-care related applications such as~\cite{boushey2017new, shao2021_ibdasystem} due to recent advances in novel computation approaches and new sensor devices. The performance of image-based dietary assessment relies on the accurate prediction of foods in the captured eating scene images. However, real world food images usually have a long-tailed distribution where a small portion of food classes (\textit{i.e.} head class) contain abundant samples for training while most food classes (\textit{i.e.} tail class) have only a few samples as shown in Figure~\ref{fig:vfn-lt}. The long-tailed classification, defined as the extreme class-imbalance problem, leads to classification bias towards head classes and poor generalization ability on recognizing tail food classes. 

As few existing long-tailed image classification methods target food images, we first introduce two benchmark long-tailed food datasets including Food101-LT and VFN-FT. Similarly as in~\cite{IMAGENET-LT}, Food101-LT is constructed as a long-tailed version of the original balanced Food101~\cite{Food-101} dataset by following the Pareto distribution. In addition, as shown in Figure~\ref{fig:vfn-lt}, VFN-LT is also used and provides a new and valuable long-tailed distributed food dataset where the number of samples for each food class exhibits the distribution of consumption frequency~\cite{lin2022differences}, defined as how often a food is consumed in one day as in the National Health and Nutrition Examination Survey\footnote[1]{https://www.cdc.gov/nchs/nhanes/index.html} (NHANES) from 2009–2016 among 17,796 U.S. healthy adults aged 18 and older, \textit{i.e. }The head classes of VFN-LT are the most frequently consumed foods in the real world for the population represented. 

An intuitive way to address the class-imbalance issue is to undersample the head classes and oversample the tail classes to obtain a balanced training set containing a similar number of samples for all classes. However, there are two major challenges: (1) How to undersample the head classes to remove the redundant samples while still keeping the original performance? (2) How to oversample the tail classes to increase the model generalization ability as the naive repeated random oversampling can further intensify the overfitting problem resulting in worse performance. 

In this work, we address both aforementioned problems by introducing a novel 2-Phase framework. Specifically,  Phase-I is the vanilla training using all images from all classes. In Phase-II, we first select the most representative data of the head classes by leveraging the model trained in phase-I as the feature extractor and then applying knowledge distillation to maintain the learned information to address issue (1). Inspired by the most recent work~\cite{CMO}, we propose a new visually-aware oversampling for the tail classes and allow multi-image mixing instead of only one image as proposed in~\cite{CMO}. The contributions of this work are summarized as following.
\begin{itemize}
    \item We introduce two benchmark datasets, Food101-LT and VFN-LT, where the long-tailed distribution of VFN-LT exhibits the real world food distribution.
    \item A novel 2-Phase framework is introduced to address the class-imbalance problem by undersampling redundant head class samples and oversampling tail classes through visual-aware multi-image mixing.
    \item We conduct extensive experiments to obtain the best classification performance on both Food101-LT and VFN-LT datasets and compare them with existing state-of-the-art methods. 
\end{itemize}

\vspace{-0.2cm}
\section{Related Work}
\vspace{-0.2cm}
\label{sec:relatedwork}
\subsection{Food Classification}
The most common methods for food classification apply off-the-shelf models such as ResNet~\cite{RESNET} to train on food image datasets~\cite{Food-101}. In addition, the prior work proposed the construction of a semantic hierarchy based on both visual similarity~\cite{mao2020visual} and nutrition contents~\cite{mao2021_nutri_hierarchy} to perform optimization on each level. Food classification has also been studied under a continual learning scenario where new foods are learned sequentially over time~\cite{He_2021_ICCVW, ILIO, he2022_expfree}. The most recent work targeted on the multi-labels ingredients recognition~\cite{gao2022dynamic_LTingredient}. The focus of this work however, is on long-tailed single food classification where each image contains only one food class and the training samples for each class are heavily imbalanced. 

\subsection{Long-tailed Classification}
Existing long-tailed classification methods can be categorized into two main groups including: (i) re-weighting and (ii) re-sampling. Re-weighting based methods aim to mitigate the class-imbalance problem by assigning tail classes or samples with higher weights than the head classes. The inverse of class frequency is widely used to generate the weights for each class as in~\cite{inverse_frequency_1,inverse_frequency_2}. In addition, a variety of loss functions are proposed to adjust the weights during training including label-distribution-aware-margin loss~\cite{LDAM}, Balanced Softmax~\cite{BSLoss} and instance based focal loss~\cite{FocalLoss}. Alternatively, re-sampling based methods aim to generate a balanced training distribution by undersampling the head classes as described in~\cite{ROS} and oversampling the tail classes as shown in~\cite{ROS, RUS_ROS}, which oversampled all tail classes until class balance was achieved. However, a drawback to undersampling is that valuable information of head classes can be lost and the naive oversampling can further intensify an overfit problem due to the lack of diversity of repeated samples. A recent work~\cite{CMO} proposes to perform oversampling by leveraging CutMix~\cite{yun2019cutmix} of head and tail classes samples. However, the performance of existing methods on food data still remain under-explored yet, which can be more challenging due to the low inter-class and high intra-class similarity. Our proposed method falls in the re-sampling category, which undersamples the head classes by selecting the most representative data while maintaining generalization ability through knowledge distillation~\cite{KD}. In addition, we propose a novel visually-aware oversampling strategy and allow multi-image CutMix compared with~\cite{CMO}. 

\section{Datasets}
\label{sec:datasets}
We introduce two benchmark datasets for long-tailed food classification including the Food101-LT and VFN-LT. 

\textbf{Food101-LT} is the long-tailed version of Food101~\cite{Food-101}, a large scale balanced dataset containing 101 food classes and 1,000 images for each class. Similar to~\cite{IMAGENET-LT}, we generate the Food101-LT by following the Pareto distribution with the power value $\alpha=6$. Overall, the training set of Food101-LT had over 11K images from 101 categories, with a maximum of 750 images per class and minimum of 5 images per class. The imbalance ratio, defined as the maximum over the minimum number of training sample, is calculated as 150. We keep the test set balanced with 250 images per class. 

\textbf{VFN-LT} is the long-tailed version of VFN~\cite{mao2020visual}, which has 74 common food classes in U.S. based on the WWEIA\footnote[2]{https://data.nal.usda.gov/dataset/what-we-eat-america-wweia-database}. However, instead of simulating the long-tail distribution as in Food101-LT, we manually matched each food class with a general food code from FNDDS\footnote[3]{https://www.ars.usda.gov/northeast-area/beltsville-md-bhnrc/beltsville-human-nutrition-research-center/food-surveys-research-group/docs/fndds-download-databases/} 2017-2018 database and assign it with the corresponding consumption frequency~\cite{lin2022differences}, which is collected through the National Health and Nutrition Examination Survey (NHANES) from 2009–2016 among 17,796 U.S. healthy adults aged 18 and older. The consumption frequency exhibits the most frequent and the least frequent consumed foods in the U.S. . Finally we randomly sample the images within each food class $i$ based on the matched consumption frequency $f_i$ as $s_i = n_i \times \frac{f_i}{f_{max}}$ where $s_i$ and $n_i$ refer to the number of selected and original data and $f_{max}$ denotes maximum matched consumption frequency in VFN. Overall, the training set of VFN-LT has 2.5K images from 74 categories, with maximum of 288 images per class and minimum of 1 image per class. The imbalance ratio is 288 and we keep the test set as balanced with 25 images per class. 

\vspace{-0.2cm}
\section{Method}
\begin{figure}[t]
\begin{center}
  \includegraphics[width=1.\linewidth]{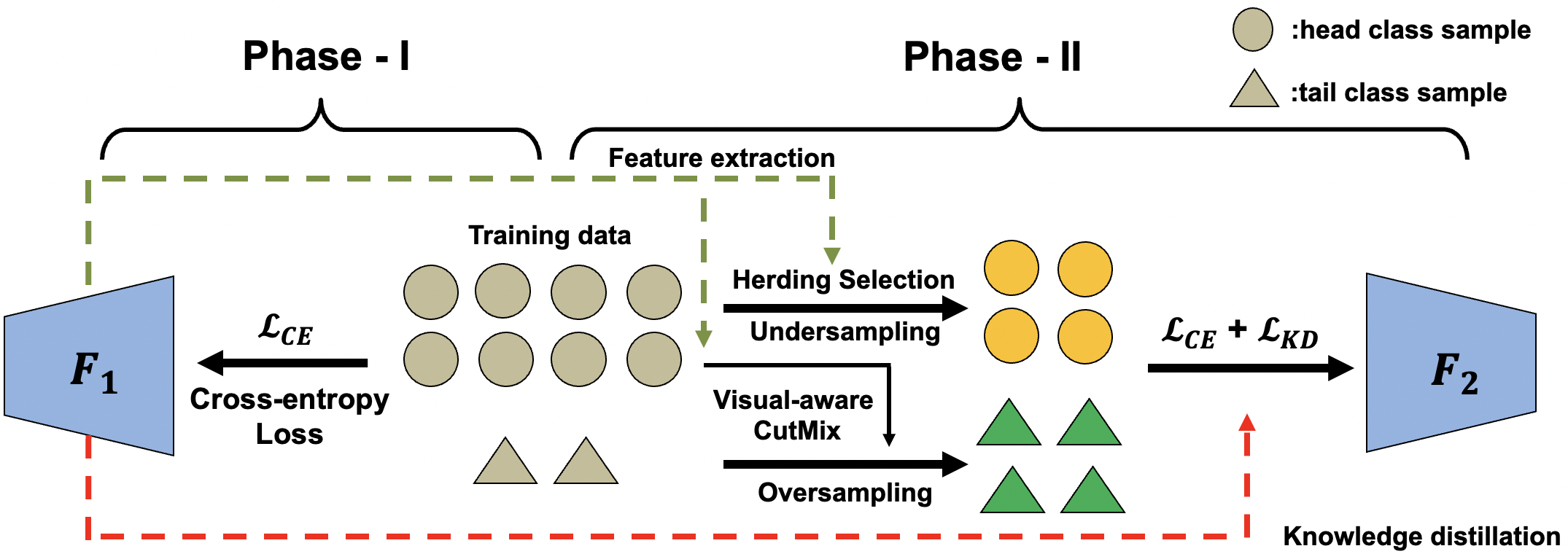}
   \vspace{-0.4cm}
  \caption{The overview of our proposed method. The left side shows the the Phase-I of vanilla training with cross-entropy loss using all the training images. The right side shows the Phase-II, where we perform undersampling for the head classes to select the most representative samples through Herding and perform oversampling for tail classes through visual-aware CutMix.}
  \label{fig:method}
\end{center}
\end{figure}

\vspace{-0.2cm}
\label{sec:method}
In this work, we introduce a novel 2-Phase framework to address the class-imbalance problem for food image classification. An overview of our proposed method is shown in Figure~\ref{fig:method} where Phase-I is the vanilla training using all the images from all classes. The trained model in Phase-I is used as the feature extractor and the teacher model for knowledge transfer in the next phase. In Phase-II, we select the most representative data from the head classes and augment the tail classes images through visual-aware CutMix~\cite{yun2019cutmix} to construct a balanced training set and apply knowledge distillation to maintain the learned information from the head classes. Details of each component are illustrated in the following subsections.

\vspace{-0.2cm}

\subsection{Undersampling and Knowledge Transfer}
\vspace{-0.2cm}
\label{subsec: undersample}
The objective is to oversample the head classes as the redundant training data can restrict the model's generalization ability~\cite{RUS_ROS}. However, the naive approach of removing a random portion of data loses much valuable information and degrades the classification performance~\cite{RUS_ROS}. We address this problem by first selecting representative samples through herding dynamic selection~\cite{HERDING} and then applying knowledge distillation to maintain the original information learned in Phase-I. Specifically, $F_1$ denotes the model trained after Phase-I by using all the training samples $D$, then we freeze the parameters of $F_1$ in Phase-II and first extract the feature embeddings using the lower layers of $F_1$ and next apply the herding algorithm to select the most representative samples for each head class based on the class mean $D_s, D_r = Herding(F_1(D))$ where $D_s$ and $D_r$ denote the selected and removed samples. Then during the Phase-II training , only selected samples from the head classes are used for training and we apply knowledge distillation~\cite{KD} to maintain the original learned knowledge as described in Equation~\ref{eq:kd}
\begin{equation} \label{eq:kd}
\begin{aligned}
\mathop{L_{KD}(x)}_{x\in D_s} = \sum_{i=1}^n -F_1^T(x)^{(i)}log[F_2^T(x)^{(i)}]
\end{aligned}
\end{equation}
where $n$ denotes the total number of classes and $F_2$ refers to the training model in Phase-II. $T>0$ is the temperature scalar in distillation where $F^T(x) = \frac{\exp{(F(x)^{(i)}}/T)}{\sum_{j=1}^n\exp{(F(x)^{(j)}}/T)}$ and we use $T=0.5$ to efficiently transfer and maintain the knowledge learned in Phase-I for the removed samples $D_r$. 

\vspace{-0.2cm}
\subsection{Oversampling of Tail Classes}
\vspace{-0.2cm}
\label{subsec: oversamples}
For the tail classes, the objective is to oversample for better generalization. The naive random oversampling~\cite{ROS} shows a severe overfitting problem. The most recent work addresses this problem by leveraging context-rich information from head classes and mixing this with samples from the tail classes through CutMix~\cite{yun2019cutmix}. However, the performance is limited when applied on food images as the selection of the head class sample is random so the mixing with visually dissimilar images can lose important semantic meaning of the original class. We address this problem by considering the visual similarity of the head class image selection and allow for up to $k$ multi-image CutMix. Specifically, during the Phase-II training, we randomly sample a head class batch $B_r \subseteq D_r$ and augment each tail class data $x\in \mathcal{R}^{W\times H \times C}$ as in Equation~\ref{eq: cutmix} where $\odot$ is the element-wise multiplication.
\begin{equation}
    \label{eq: cutmix}
    \begin{aligned}
    \Tilde{x} = \sum_{i=1}^k M^s \odot x_k + (1-M^s) \odot x
    \end{aligned}
\end{equation}
$M^s\in\{0,1\}^{sW\times sH}$ refers to the binary mask indicating where to cut and paste for the two images and $0<s<1$ is the randomly sampled mixing ratio. $x_1, x_2,...x_k$ denote the top-k most similar head class images with highest cosine similarity $cos$ calculated by
\begin{equation}
    \label{eq:cosine}
    \begin{aligned}
   x_1, x_2,..., x_k = \mathop{\textit{argmax}}_{x_r \in B_r} cos(F_1(x), F_1(x_r))
    \end{aligned}
\end{equation}
We set $k=1$ for our method and observe improved performance by slightly increasing $k$, but very large $k$ can harm the performance since the distribution of augmented data becomes different from the original as shown in Section~\ref{subsec: ablation study}.

\section{Experiments}
\vspace{-0.2cm}
\label{sec:exp}
\subsection{Experimental Setup}
\vspace{-0.2cm}
\label{subsec:exp setup}
\textbf{Datasets and evaluation metrics.} We validate our method on two benchmark datasets introduced in Section~\ref{sec:datasets} including Food101-LT and VFN-LT. For each dataset, we calculate the mean number of samples by $m = \frac{D}{n}$ where $D$ and $n$ denote the total number of samples and classes, respectively. We perform undersampling of head classes (containing more than $m$ samples) and oversampling of tail classes (lower than $m$) to achieved balanced $m$ samples per class. Overall, Food101-LT has 101 classes including 28 head classes and 73 tail classes. VFN-FT has 74 classes including 22 head classes and 52 tail classes. Top-1 classification accuracy is reported along with the accuracy for head classes and tail classes, respectively.\newline
\textbf{Comparison methods.} We compare our method with existing approaches for food classification and long-tailed classification including (1) Vanilla training as \textbf{Baseline} (2) Hierarchy based food classification (\textbf{HFR})~\cite{mao2020visual} (3) Random oversampling (\textbf{ROS})~\cite{ROS} (4) Random undersampling (\textbf{RUS})~\cite{RUS_ROS} (5) Context-rich Oversampling (\textbf{CMO})~\cite{CMO} (6) Label-distribution-aware-margin (\textbf{LDAM})~\cite{LDAM} (7) Balanced Softmax (\textbf{BS})~\cite{BSLoss} (8) Influence-balanced loss (\textbf{IB})~\cite{IBLoss} and (9) \textbf{Focal} loss~\cite{FocalLoss}.\newline
\textbf{Implementation details.} For both datasets, we use ResNet-18~\cite{RESNET} and train for 150 epochs. We apply SGD optimizer with momentum of 0.9 and the initial learning rate is 0.1 with a cosine decay schedule. For our 2-Phase method, we use 50 epochs for Phase-I and 100 epochs for Phase-II. We ran all experiments 5 times and report the average performance. 

\vspace{-0.4cm}

\subsection{Experimental Results}
\vspace{-0.1cm}
\label{subsec: exp results}
The results on Food101-LT and VFN-LT are summarized in Table~\ref{tab:expresult}. We observe poor generalization ability for tail classes with very low tail accuracy in the baseline and HFR~\cite{mao2020visual} due to the limited number of training samples. Although the naive random undersampling~\cite{ROS,RUS_ROS} increases the tail accuracy due to the decrease of class-imbalance level, the performance on the head classes drops a lot and the overall accuracy remain almost unchanged. In addition, all existing long-tailed classification methods~\cite{LDAM,BSLoss,IBLoss,CMO,FocalLoss} show improvements, but the performance is still limited as food image classification is more challenging. Our proposed method achieves the best performance on both datasets with competitive head class accuracy using only part of the training samples and achieving much higher accuracy for the tail classes. Results show the effectiveness of our undersampling strategy along with knowledge distillation to maintain the learned information for head classes and the use of visually-aware augmentation for better generalization on the tail classes. 
\begin{table}[t!]
    \centering
    \scalebox{.85}{
    \begin{tabular}{|c|ccc|ccc|}
        \hline
        Datasets & \multicolumn{3}{c|}{\textbf{Food101-LT}} & \multicolumn{3}{c|}{\textbf{VFN-LT}} \\
        \hline
        Accuracy(\%) & Head & Tail & Overall & Head & Tail & Overall \\
        \hline
        Baseline &65.8 & 20.9 & 33.4 & \textbf{62.3} & 24.4 & 35.8  \\
        HFR~\cite{mao2020visual}&\textbf{65.9} & 21.2 & 33.7 & 62.2 & 25.1 & 36.4  \\
        ROS~\cite{ROS} & 65.3 & 20.6 & 33.2 & 61.7 & 24.9 & 35.9 \\
        RUS~\cite{RUS_ROS} & 57.8 & 23.5 & 33.1 & 54.6 & 26.3 & 34.8 \\
        CMO~\cite{CMO} & 64.2 & 31.8 & 40.9 & 60.8 & 33.6 &42.1\\
        LDAM~\cite{LDAM} & 63.7 & 29.6 & 39.2 & 60.4 & 29.7 & 38.9 \\
        BS~\cite{BSLoss} & 63.9& 32.2 & 41.1 & 61.3 & 32.9 & 41.9 \\
        IB~\cite{IBLoss} & 64.1 & 30.2 & 39.7 & 60.2 & 30.8 & 39.6 \\
        Focal~\cite{FocalLoss} & 63.9 & 25.8 & 36.5 & 60.1 & 28.3 & 37.8 \\
        \hline
        Ours & 65.2 & \textbf{33.9} & \textbf{42.6} & 61.9 & \textbf{37.8} & \textbf{45.1} \\       
        \hline
    \end{tabular}
    }
    \vspace{-0.2cm}
    \caption{Top-1 accuracy on Food101-LT and VFN-LT with tail classes (Tail) and head classes (Head) accuracy.  }
    \label{tab:expresult}
\end{table}


\vspace{-0.3cm}
\subsection{Ablation Study}
\vspace{-0.2cm}
\label{subsec: ablation study}
In this section, we first evaluate our head class undersampling by comparing RUS~\cite{RUS_ROS} with (i) replacing with Herding selection (\textbf{HUS}) and (ii) applying knowledge distillation (\textbf{HUS+KD}). Then we evaluate tail class oversampling by comparing CMO~\cite{CMO} with (i) considering visual similarity (\textbf{CMO+Visual}) and (ii) increasing the number of mixed images $k$ as described in Section~\ref{subsec: oversamples}.
The results are summarized in Table~\ref{tab:ablationstudy}. The herding selection show better performance compared with random sampling as we maintain the most representative data for the head classes. Applying knowledge distillation further improves the performance without compromising performance for the head classes. In addition, we observe improvements on the tail classes when performing CutMix on visually similar food images, which avoids losing important semantic information while maintaining discriminative ability. Finally, we show better generalization ability on the tail classes by slightly increasing $k$ while the performance drops for very large $k$ due to the distribution drift of the augmented tail class images. 

\begin{table}[h]
    \centering
    \scalebox{.7}{
    \begin{tabular}{|c|ccc|ccc|}
        \hline
        Datasets & \multicolumn{3}{c|}{\textbf{Food101-LT}} & \multicolumn{3}{c|}{\textbf{VFN-LT}} \\
        \hline
        Accuracy(\%) & Head & Tail & Overall & Head & Tail & Overall \\
        \hline
        RUS~\cite{RUS_ROS} & 57.8 & 23.5 & 33.1 & 54.6 & 26.3 & 34.8 \\
        HUS & +1.7 & +0.3 & +0.6 & +2.1 & +0.2 & +0.7 \\
        HUS+KD & +5.8 & +0.2 & +1.9 & +7.1 & +0.1 & +2.2 \\
        \hline
        CMO~\cite{CMO} & 64.2 & 31.8 & 40.9 & 60.8 & 33.6 &42.1\\
        CMO+Visual (k=1) & +0.2 & +1.3 & +1.0 & +0.4 & +1.8 & +1.2\\
        CMO+Visual (k=3) & +0.1 & +2.8 & +2.1 & +0.5 & +2.4 & +1.9\\
        CMO+Visual (k=10) & -0.1 & +0.2 & +0.1 & +0.2 & -0.4 & -0.1\\
        \hline
    \end{tabular}
    }
    \vspace{-0.2cm}
    \caption{Ablation study on Food101-LT and VFN-LT.}
    \label{tab:ablationstudy}
\end{table}

\vspace{-0.5cm}
\section{Conclusion and Future Work}
\vspace{-0.3cm}
\label{sec:conclusion}
In this work, we focused on the long-tailed data distribution problem for food image classification. 
We introduced two new benchmark long-tailed datasets, Food101-LT and VFN-LT, where the VFN-LT exhibits real world distribution of foods with severe class-imbalance. We introduced a novel 2-Phase framework to select the most representative training samples for head classes while maintaining the information through knowledge distillation, and augment the tail classes by visually-aware multi-image CutMix. Our method achieved the best performance on both datasets and extensive experiments were conducted to evaluate the contribution of each component in proposed method. Our future work will design single-phase, end-to-end food classification system. 

\bibliographystyle{IEEEbib}
{\small\bibliography{strings,refs}}

\end{document}